\documentclass[letterpaper, 10 pt, conference]{ieeeconf}  

\IEEEoverridecommandlockouts  

\usepackage{times}

\usepackage{cite}
\usepackage{multicol}
\usepackage{graphics} 
\usepackage{amsmath} 
\usepackage{amssymb}  
\usepackage{algorithm}
\usepackage{multirow}
\usepackage{algpseudocode}
\usepackage{subcaption}
\usepackage[dvipsnames]{xcolor}
\usepackage{hyperref}
\hypersetup{
    colorlinks=true,
    linkcolor=black,    
    urlcolor=cyan,
    citecolor=black
    }

\usepackage{epsfig} 
\usepackage{times} 
\usepackage{amsmath} 
\usepackage{amssymb}  

\usepackage{algpseudocode}

\newcommand{\ours}{\textbf{IMRL}}

\title{\Large \bf IMRL: Integrating Visual, Physical, Temporal, and Geometric Representations for Enhanced Food Acquisition}

\author{Rui Liu, Zahiruddin Mahammad, Amisha Bhaskar, Pratap Tokekar   
\thanks{All authors are from the University of Maryland, College Park, MD 20742, USA. Email: \tt\small \{ruiliu, zahirmd, amishab, tokekar\}@umd.edu}
\thanks{This work is supported in part by an Amazon Research Award.}
}

\begin{document}

\maketitle

\begin{abstract}
Robotic assistive feeding holds significant promise for improving the quality of life for individuals with eating disabilities. However, acquiring diverse food items under varying conditions and generalizing to unseen food presents unique challenges. Existing methods that rely on surface-level geometric information (e.g., bounding box and pose) derived from visual cues (e.g., color, shape, and texture) often lacks adaptability and robustness, especially when foods share similar physical properties but differ in visual appearance. We employ imitation learning (IL) to learn a policy for food acquisition. Existing methods employ IL or Reinforcement Learning (RL) to learn a policy based on off-the-shelf image encoders such as ResNet-50. However, such representations are not robust and struggle to generalize across diverse acquisition scenarios. To address these limitations, we propose a novel approach, IMRL (Integrated Multi-Dimensional Representation Learning), which integrates visual, physical, temporal, and geometric representations to enhance the robustness and generalizability of IL for food acquisition. Our approach captures food types and physical properties (e.g., solid, semi-solid, granular, liquid, and mixture), models temporal dynamics of acquisition actions, and introduces geometric information to determine optimal scooping points and assess bowl fullness. IMRL enables IL to adaptively adjust scooping strategies based on context, improving the robot's capability to handle diverse food acquisition scenarios. Experiments on a real robot demonstrate our approach's robustness and adaptability across various foods and bowl configurations, including zero-shot generalization to unseen settings. Our approach achieves an improvement up to $35\%$ in success rate compared with the best-performing baseline. More details can be found on our website \href{https://ruiiu.github.io/imrl}{https://ruiiu.github.io/imrl}.
\end{abstract}

\begin{figure}[t]
    \centering
    \includegraphics[width=\linewidth]{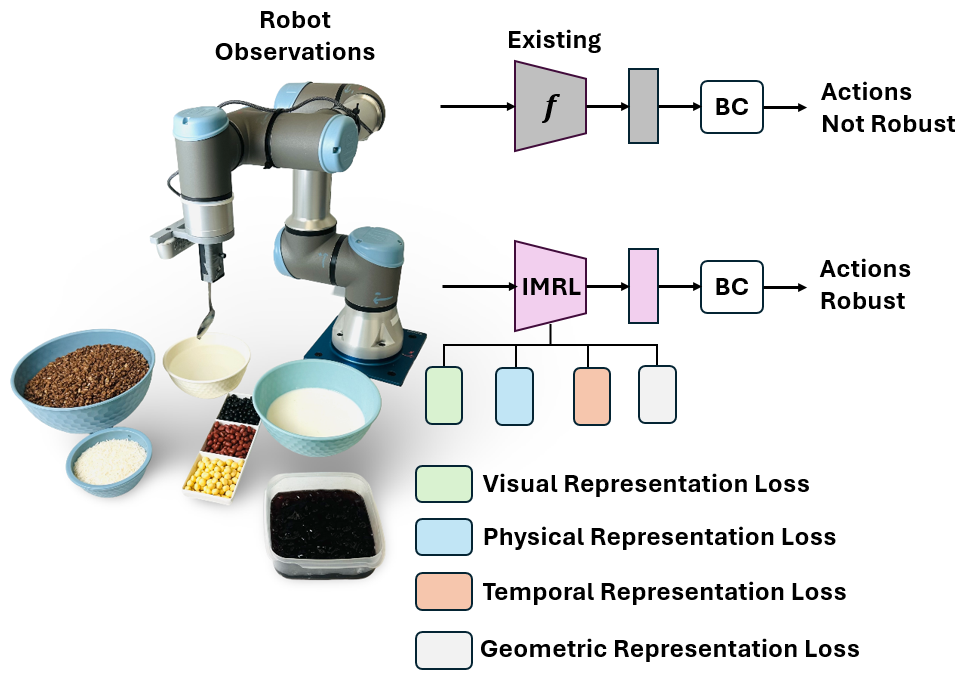}
    \caption{\textbf{Comparison of standard BC and our approach for food acquisition.} The standard BC (top) processes robot observations through an off-the-shelf encoder (e.g., ResNet-50), generating actions that are not robust and generalizable. In contrast, our proposed approach \ours~(bottom) utilizes visual, physical, temporal and geometric representation learning to develop richer and more informative representations to enhance the robustness and generalizability of BC for food acquisition.}
    \label{fig:comp}
    \vspace{-10pt}
\end{figure}
 
\section{Introduction}
Robotic assistive feeding \cite{park2020active, ohshima2013meal, Meet_obi, chokemealtime, feng2019robot, gordon2020adaptive, gordon2021leveraging, grannen2022learning}, involves the use of robotic systems to assist individuals with feeding disabilities to eat, presents a great opportunity to enhance the quality of life by providing autonomy in daily eating tasks. One of the critical challenges in this domain is the acquisition of diverse food items from various containers under different conditions. Unlike robotic manipulation tasks \cite{kopicki2011learning, zeng2021transporter, harada2012pick} that often involve rigid, well-defined objects, food acquisition introduces complexities such as variable food types, non-rigid and deformable food, different physical properties (e.g., solid, semi-solid, granular, liquid, and mixture), and different containers. These factors complicate the development of effective robotic feeding systems.

Despite progress in food acquisition for robotic assistive feeding, several limitations persist. Some systems \cite{park2020active, ohshima2013meal, Meet_obi, chokemealtime} rely on predefined, heuristic-based, or hard-coded primitive actions that lack generalization across different food types and configurations. Other works \cite{feng2019robot, gordon2020adaptive, gordon2021leveraging, sundaresan2022learning, grannen2022learning, liuadaptive, bhaskar2024lava} integrate visual perception to handle food acquisition but primarily focus on extracting information such as bounding boxes and food poses from visual cues. Capturing the food’s physical properties, crucial for effective manipulation, often requires haptic sensors, as seen in \cite{gordon2021leveraging, sundaresan2022learning}. Additionally, the acquisition of semi-solid and liquid foods remains relatively underexplored in assistive feeding, with most research focusing on solid or granular foods \cite{feng2019robot, tai2023scone, grannen2022learning}. 

We employ imitation learning (IL) and use Behavior Cloning (BC) as the IL algorithm to learn a policy for food acquisition. BC has been employed in previous works \cite{tai2023scone, liuadaptive, bhaskar2024lava} for similar tasks. However, standard BC often lacks robustness and struggles to generalize across diverse acquisition scenarios. The effectiveness of BC is heavily dependent on the quality of the representations used. Although we only focus on BC, since our advancement is in the representations, we expect that this will be just as useful in other downstream approaches such as RL \cite{hu2023imitation, bhaskar2024navinact}.

To address these gaps, our approach focuses on enhancing the robustness and generalizability of BC by developing richer and more informative representations for food acquisition in assistive feeding. We propose a novel approach that integrates visual, physical, temporal, and geometric representations to achieve that. As illustrated in Fig. \ref{fig:comp}, our pipeline incorporates visual and physical representations to capture both food types and their physical properties. The core idea is to develop representations that not only distinguish between different food types but also encode their physical characteristics. This is crucial for downstream BC tasks, as foods with similar physical properties but different appearances (e.g., water and milk) may necessitate similar scooping strategies, while foods with the same color but different physical properties (e.g., milk and rice) may require distinct strategies. Unlike prior works \cite{gordon2021leveraging, sundaresan2022learning} that rely on explicit haptic sensors to collect physical data, our approach learns an a priori representation to infer food physical properties just from visual data.

Additionally, by integrating temporal representation, we capture the temporal dynamics of food acquisition, which enhances the decision-making process over time. Leveraging geometric representation, we extract information such as optimal scooping points and depth cues like the fullness of the bowl. These types of information are vital for effective food manipulation: optimal scooping points guide the model on where to scoop the desired food, while understanding the fullness of the bowl allows the robot to adapt its scooping strategy—requiring deeper scooping as the food quantity decreases.


To summarize, the key contributions of our work are as follows:
\begin{itemize}
    \item We introduce a novel approach \ours~that integrates visual, physical, temporal, and geometric representations to provide a richer understanding of foods. This advanced representation captures details that are critical for effective food acquisition, going beyond surface-level visual information.
    
    \item Our approach improves the robustness and generalizability of BC by leveraging these enhanced representations. This advancement ensures that the learned policies can effectively handle a wide range of food types and configurations, including unseen scenarios, addressing the limitations of standard BC methods. \ours~achieves up to a $35\%$ improvement in success rate compared to the best-performing baseline in testing with a real UR3 robot.
\end{itemize}

\section{Related Work} \label{sec:related}


\paragraph*{Food Acquisition for Assistive Feeding}
While recent studies have explored various methods for food acquisition for assistive feeding, achieving robust and adaptive food acquisition is still challenging. Several commercially available devices for mealtime assistance such as \cite{Meet_obi, chokemealtime}, often rely on pre-defined trajectories or require user teleoperation, limiting their adaptability. Feng et al. \cite{feng2019robot}, Gordon et al. \cite{gordon2020adaptive, gordon2021leveraging} and Sundaresan et al. \cite{sundaresan2022learning} leveraged visual information to plan fork skewering motions. Similarly, Grannen et al. \cite{grannen2022learning} planned bimanual food scooping and grouping actions. However, these approaches primarily focus on learning geometric information (e.g., bounding box, food pose) from visual cues for food acquisition; they do not capture the physical properties of the food, or they rely on haptic sensors to determine the necessary force for interacting with food. Tai et al. \cite{tai2023scone} introduced a food scooping framework, but their work focused solely on granular foods like beans and rice, neglecting other food types such as semi-solids and liquids, and did not account for variations in bowl configurations (e.g., different sizes or shapes).

\paragraph*{Visual Representation Learning}
Visual Representation learning has been explored in the context of robotic manipulation and imitation learning, where visual features are crucial for understanding object properties and guiding actions. Previous works such as Dense Object Nets (DON) \cite{florence2018dense} focus on learning dense visual object descriptors for robotic manipulation by leveraging 3D information, including depth data, to capture fine-grained object details and correspondences. However, these approaches either assume the availability of depth data or depend on substantial supervision to capture the object's state and properties effectively. By using exclusively RGB data, we avoid the complexities associated with 3D reconstruction and depth data while still achieving robust and informative representations that are applicable to downstream imitation learning tasks.

Other methods, like DINO-v2 \cite{oquab2023dinov2}, SwAV \cite{caron2020unsupervised}, MoCo-v3 \cite{chen2021empirical}, and SimSiam \cite{chen2021exploring} utilize self-supervised or unsupervised learning to learn representations. However, representations learned by these methods often focus on surface-level features such as color, shape, and texture, and have limited understanding of the physical properties of objects.  For example, a pretrained ResNet-50 \cite{he2016deep} can distinguish between water and milk based on their color differences but lacks the understanding that both are liquids. Similarly, it might recognize that milk and yogurt are both white, but it does not comprehend that one is a liquid while the other is semi-solid. In contrast, our approach learns not only to classify food types, but also learns the physical properties of food (e.g., liquid, solid, granular, semi-solid, and mixture), as well as captures temporal dynamics for food acquisition tasks, distinguishing it from other methods that focus mainly on visual surface features. 

\paragraph*{Imitation Learning}
Imitation learning \cite{fang2019survey, hussein2017imitation, ARGALL2009469} has been widely utilized in robotic manipulation to enable robots to mimic human demonstrations and perform complex tasks. Traditional approaches like Behavior Cloning (BC) \cite{florence2019self} focus on learning a policy that maps observations directly to actions by minimizing the error between the predicted and demonstrated actions. While effective, standard BC often lacks robustness and generalizability across diverse contexts, as its performance is highly reliant on the quality of the learned representations. Our approach addresses these limitations by developing richer and more informative representations, which include awareness of food types, physical properties, temporal dynamics, optimal scooping points, and bowl fullness. This allows the robot to adapt its actions based on the specific attributes of the food and the environment, resulting in more robust and generalizable BC for food acquisition.


\begin{figure}[t]
    \centering
    \includegraphics[width=\linewidth]{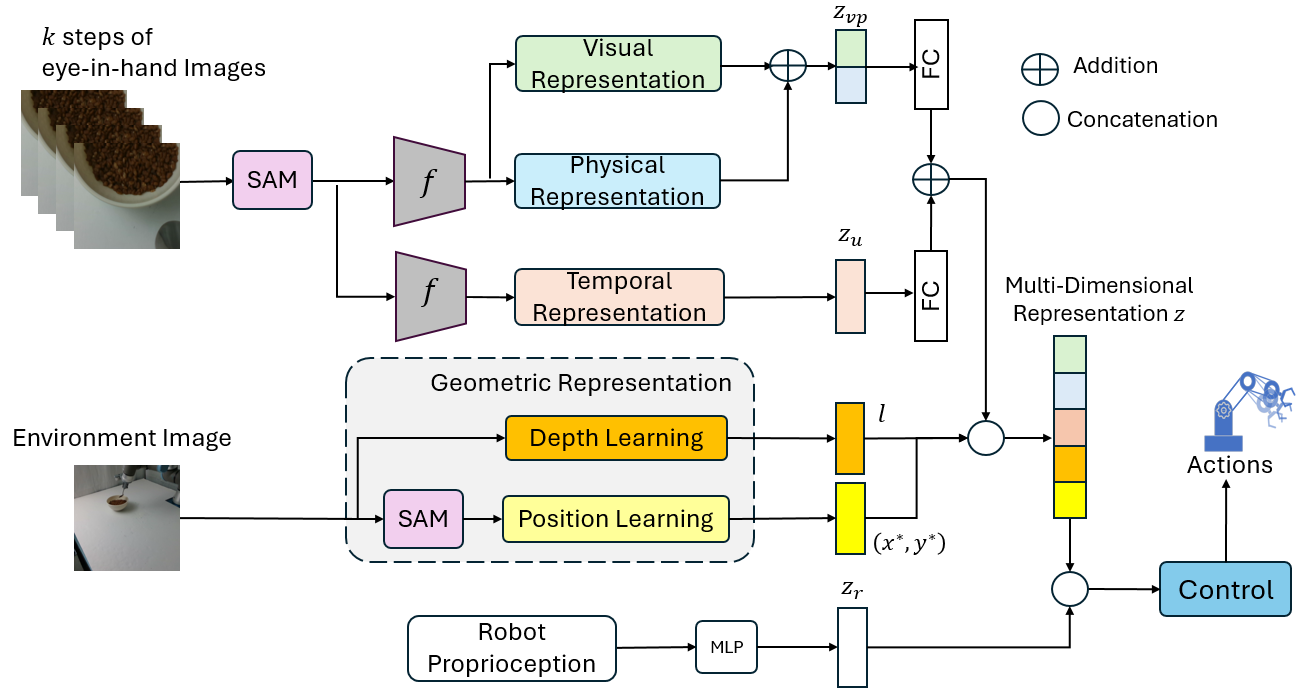}
    \caption{\textbf{Overview of the proposed IMRL approach for food acquisition}. Given the last $k$ steps of eye-in-hand RGB observations, the system segments the desired food using SAM, extracts features with an encoder (e.g., ResNet-50), and processes them through visual and physical representation modules to learn a joint representation $z_{vp}$. A temporal representation module produces $z_u$ to capture dynamics, while the geometric representation module provides bowl fullness $l$ and optimal scooping points $(x^*, y^*)$. All these representations are integrated into a multi-dimensional representation $z$, which, combined with robot proprioception, is used to generate robot actions through a control module.}
    \label{fig:app}
    \vspace{-10pt}
\end{figure}

\vspace{-5pt}
\section{Approach} \label{sec:approach}
\subsection{Problem Formulation}
We formulate the food acquisition task as a policy learning problem to learn a visuomotor policy $\pi_\theta$, parameterized by $\theta$. The input observation space is defined as $\mathcal{O}_t = (\mathcal{I}_t, p_t)$, where $\mathcal{I}_t \in \mathbb{R}^{3 \times H \times W}$ represents the RGB images, and $p_t \in \mathbb{R}^6$ denotes the robot proprioception, representing the 6D joint angles. The state $s_{t-k:t} = (\mathcal{I}_{t}^e, \mathcal{I}_{t-k:t}^h, p_{t-k:t})$ includes an environment RGB image and a sequence of the last $k$ steps of eye-in-hand RGB images and joint angles up to the current timestamp $t$. The output action consists of the predicted joint angles for the next step.

\subsection{Representation Learning} \label{sec:vrl}
Representation learning is essential as it provides compact, rich representation that is critical for downstream tasks such as BC, directly influencing BC's performance. We begin by adding image data augmentation, including random center crop, horizontal flip, color jittering, and Gaussian blur. Next, we use the Segment Anything Model (SAM) \cite{kirillov2023segment} to segment the desired food. We then extract features from a pretrained model, such as ResNet-50 \cite{he2016deep}, pretrained on ImageNet \cite{deng2009imagenet}.

After extracting features, it is essential to fine-tune these features to make them aware of not only food types but also physical properties to improve performance for our downstream BC tasks. To achieve this, we collect a dataset of food images featuring various types with different physical properties, including liquids (e.g., milk and water), granular substances (e.g., rice and cereals), semi-solids (e.g., jello and yogurt), and mixture items (e.g., soup with both liquid and solid components).

\begin{figure}[t]
    \centering
    \includegraphics[width=\linewidth]{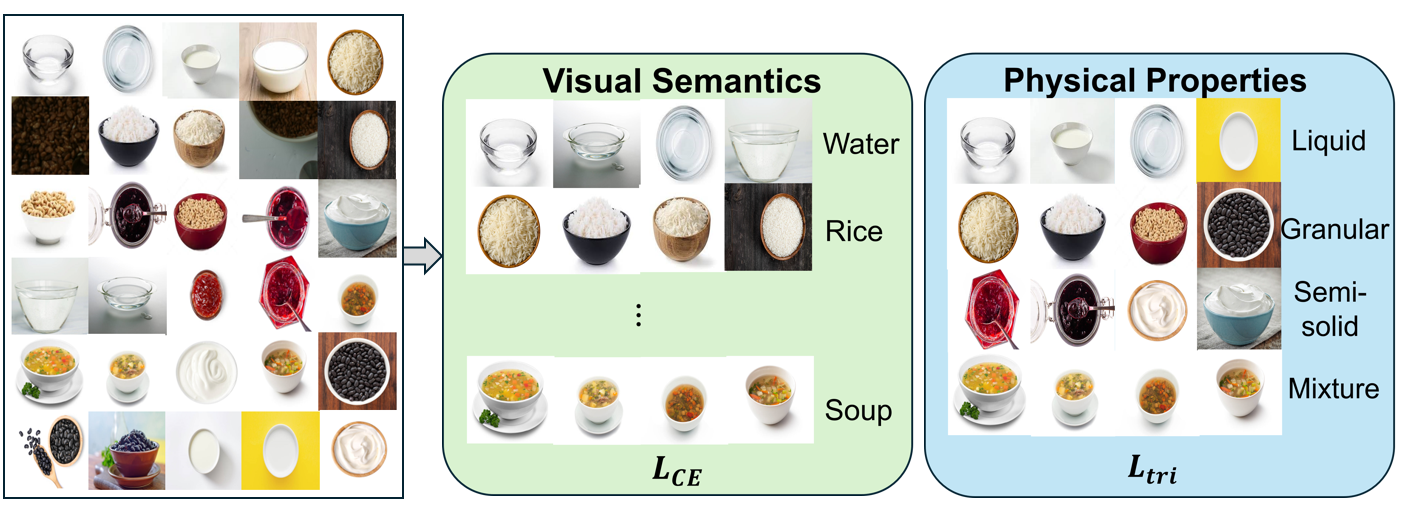}
    \caption{\textbf{Visual and physical representation learning.} Given a food image dataset, we not only learn visual semantics of different food types with a cross-entropy loss, but also food physical properties with a triplet loss.}
    \label{fig:vpr}
    \vspace{-15pt}
\end{figure}


\subsubsection{Visual Semantics}
To make the features visually food type aware and encourage the model to learn better visual semantic representations, as shown in Fig. \ref{fig:vpr}, we finetune the model by minimizing the following cross-entropy loss: $\mathcal{L}_{CE} = -\mathbb{E}_D \big[\sum_{i=1}^C y_i \log(p_i)\big]$, 
where $C$ is the total number of classes in the food dataset, $y_i$ is the ground truth label, and $p_i$ is the predicted probability of an image $x$ belongs to class $i$. 

\subsubsection{Physical Properties}
As previously mentioned, understanding the physical properties of food is crucial. Therefore, instead of just learning visual semantics, we also focus on learning the physical properties of food, as illustrated in Fig. \ref{fig:vpr}. The goal is to develop representations that not only differentiate food types but also capture physical properties. Foods with similar physical properties should have closer representations, while those with different properties should be more distinct. For example, milk and water, which are both liquids, should have closer representations. In contrast, milk and rice should have more distinct representations despite both being white, and rice and cereals should be closer due to their shared granular property.

We apply contrastive learning with triplet loss to ensure that representations of food items with similar physical properties are pulled closer together, while those with differing properties are pushed further apart. We define a triplet as following: $\mathcal{A}$ (anchor), a sample whose representation that we want to anchor (e.g., milk); $\mathcal{P}$ (positive), a sample that is similar to the anchor in terms of the physical property (e.g., water, which shares the liquid property with milk); $\mathcal{N}$ (negative), a sample that is different from the anchor (e.g., rice, which is granular). We use the L2 norm to measure the dissimilarity between feature representations: $d(f(x_i), f(x_j)) = ||f(x_i)-f(x_j)||_2$,
where $f$ is the feature extractor that maps an input image $x$ to its feature representation $f(x)$. Mathematically, we define the triplet loss for a triplet $(\mathcal{A}, \mathcal{P}, \mathcal{N})$ as: $\mathcal{L}(\mathcal{A}, \mathcal{P}, \mathcal{N}) = \max \big(0, d(f(\mathcal{A}), f(\mathcal{P}))-d(f(\mathcal{A}), f(\mathcal{N}))+\alpha \big)$, 
where $\alpha$ is the margin, a hyperparameter that defines how much closer the positive should be compared to the negative. The objective is to make the distance between the anchor and positive $d(f(\mathcal{A}),f(\mathcal{P}))$ smaller than the distance between the anchor and negative $d(f(\mathcal{A}),f(\mathcal{N}))$ by at least a margin $\alpha$. The overall triplet loss is: $\mathcal{L}_{tri} = \mathbb{E}_D [\mathcal{L}(\mathcal{A}, \mathcal{P}, \mathcal{N})]$.


\begin{figure*}[ht]
     \centering
     \begin{subfigure}[b]{0.325\textwidth}
         \centering
         \includegraphics[width=\textwidth]{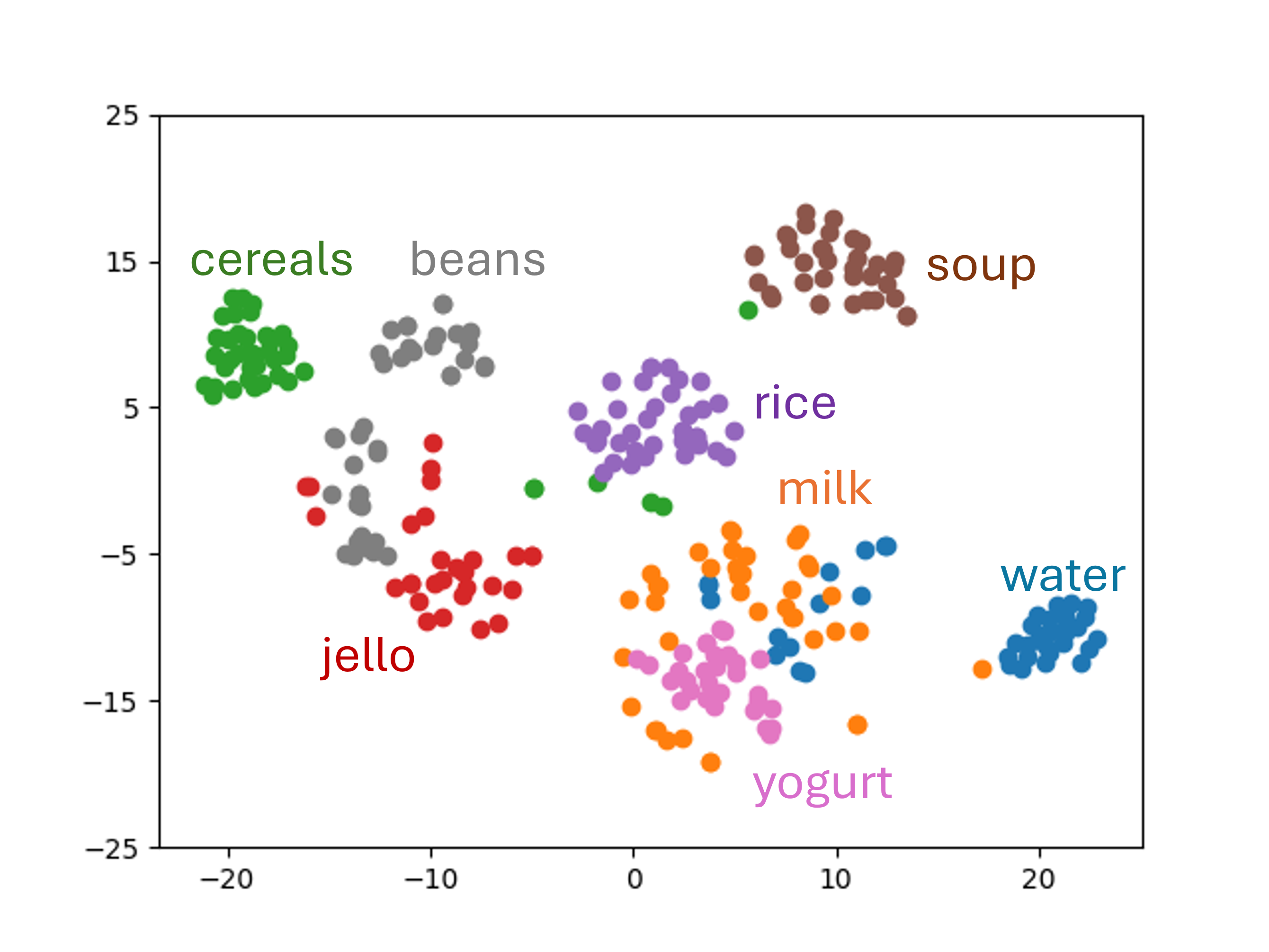}
         \caption{Pretrained ResNet-50 on ImageNet}
         \label{fig:pretrained}
     \end{subfigure}
     \begin{subfigure}[b]{0.325\textwidth}
         \centering
         \includegraphics[width=\textwidth]{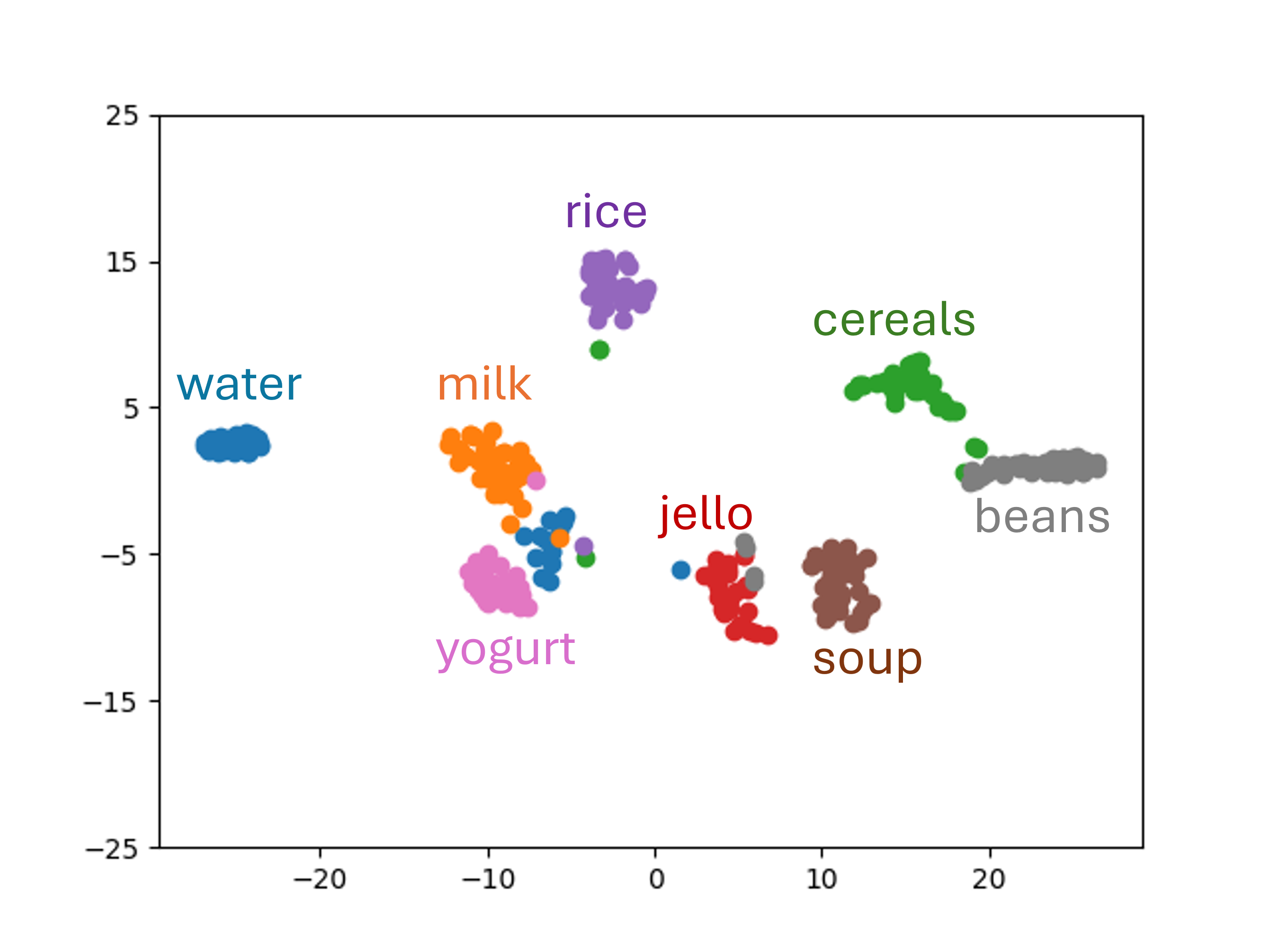}
         \caption{Visual semantics only}
         \label{fig:visual_only}
     \end{subfigure}
     \begin{subfigure}[b]{0.325\textwidth}
         \centering
         \includegraphics[width=\textwidth]{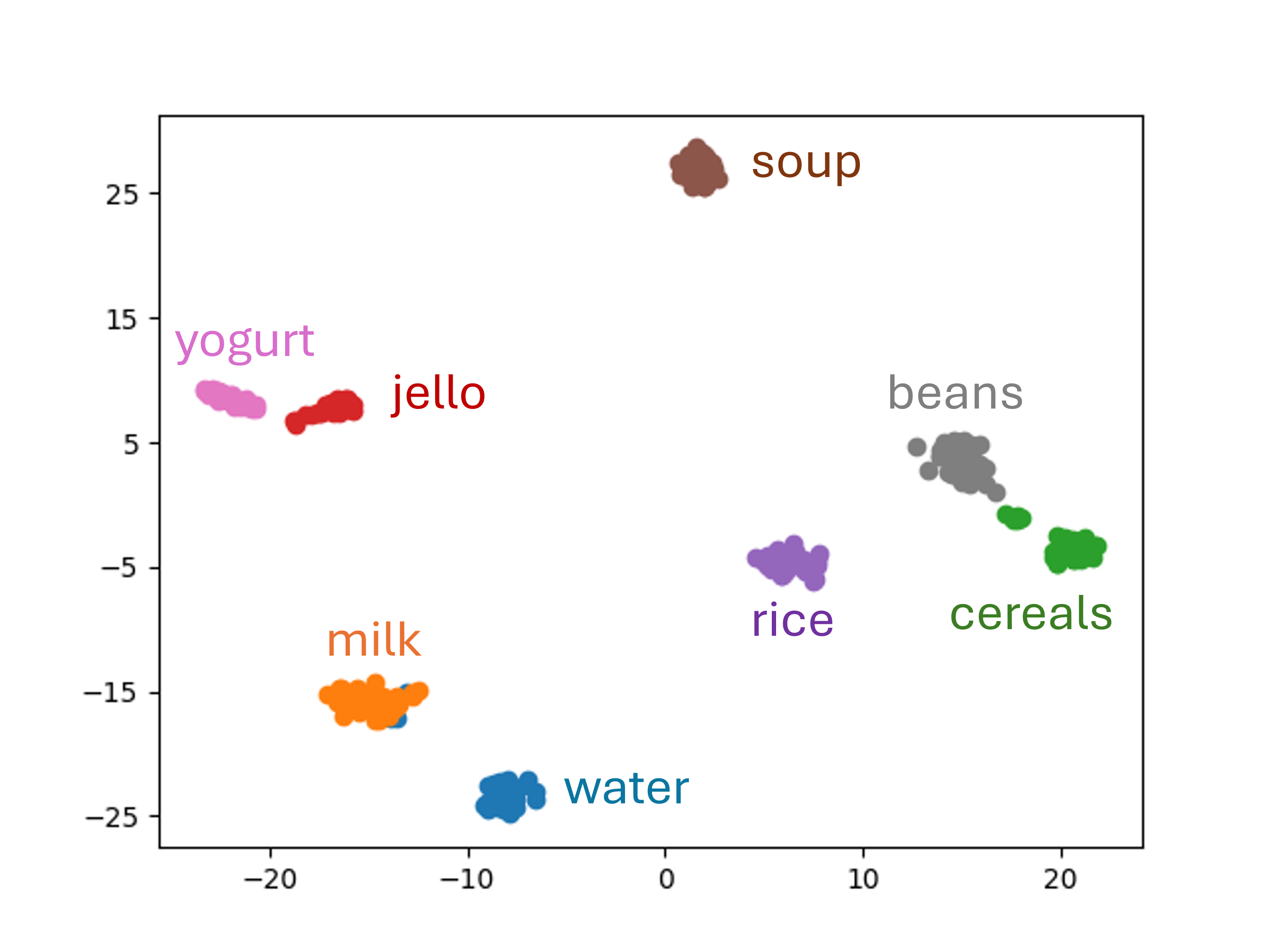}
         \caption{Visual semantics and physical properties}
         \label{fig:vp}
     \end{subfigure}
        \caption{\textbf{2D t-SNE visualization of representation embeddings of different food items.} The embeddings are generated using (a) a pretrained ResNet-50 on ImageNet, (b) visual representation learning considering visual semantics only, and (c) visual and physical representation incorporating both visual semantics and physical properties of the food.}
        \label{fig:embed}
        \vspace{-15pt}
\end{figure*}

\subsubsection{Temporal Dynamics} \label{sec:tdl}
While visual and physical representation provides features that capture food types and physical properties, which is advantageous for downstream BC tasks. However, food acquisition tasks involve predicting future actions based on current and past observations, making them sensitive to temporal dynamics. 


Given video sequences with frames $\{x_1, x_2, \ldots, x_N \}$ of the robot scooping food from bowls, we shuffle these frames and then aim to predict the original order for each frame $x_j$. We formulate this temporal dynamics learning task as a classification problem with the following cross entropy loss: $\mathcal{L}_{temp} = -\mathbb{E}_D \big[\sum_{j=1}^N y_j\log(p_j) \big]$, 
where $N$ is the number of frames in a video sequence, $y_j$ is the ground truth label for the correct order of frame $x_j$, and $p_j$ is the predicted probability that the frame $x_j$ is in position $j$. This loss to predict the original order of shuffled video frames helps learn temporal dynamics by ensuring that the model captures and understands the sequential relationships between frames.

\subsubsection{Geometric Information} \label{sec:gl}
Visual, physical, and temporal representations capture food types, properties, and dynamics but miss critical details like food position and bowl fullness. Incorporating geometric information, such as position and depth, addresses these gaps: position-aware learning identifies optimal scooping points, depth-aware learning adjusts the scooping strategy based on bowl fullness. Integrating these information allows the robot to adapt its scooping strategy to the food’s location and the bowl’s fullness, improving its ability to scoop and clear the bowl effectively.



\textbf{Position-Aware Learning.}
Position information enables the robot to focus on specific food items and determine effective scooping strategies. To identify the optimal scooping points, we propose the \textbf{interior point search with local density} method to determine optimal scooping points within segmented food regions. This method leverages local density information to identify key points that are not only within the food area but also avoid sparse regions and minimize collisions with the bowl. 

We first use SAM \cite{kirillov2023segment} to segment the desired food item and generate a binary mask over the food. SAM is known for its high-quality segmentation and adaptability without retraining. We then compute the expectation over the mask to determine the centroid position of the food, which serves as a reference point to initialize the search for the optimal scooping point. To find the optimal scooping point, we calculate local density for each point $(x_i, y_i)$ inside the food mask $(M(x_i, y_i)=1)$, we define a local neighborhood $N_r(x_i, y_i)$ of size $r\times r$ centered around $(x_i, y_i)$. The local density $\rho(x_i, y_i)$ for that point is given by: $\rho(x_i, y_i) = \frac{\sum_{(x_j, y_j)\in N_r(x_i, y_i)} M(x_j,y_j)}{|N_r(x_i, y_i)|}$,
where $|N_r(x_i, y_i)|$ is the total number of pixels in the neighborhood $N_r(x_i, y_i)$. This equation computes the proportion of food pixels in the local region, effectively providing a measure of food density around point $(x_i, y_i)$. The optimal scooping point $(x^*, y^*)$ is the point inside the mask that maximizes the local density: $(x^*, y^*) = \arg \max_{(x_i,y_i)} \{\rho(x_i,y_i) | M(x_i, y_i) = 1\}$. 

We also ensure that the chosen point $(x^*, y^*)$ lines within the interior of the mask. We achieve this by defining a condition that points selected are at least a margin distance $m$ away from the mask boundary. This constraint also helps avoid collision with the bowl's edges when scooping: $\text{min distance}\big((x^*, y^*), \mathcal{B}(M)\big) > m$, 
where $\mathcal{B}(M)$ represents the boundary of the mask $M$. 

\textbf{Depth-Aware Learning.}
To estimate the fullness of the bowl, we employ regression with Mean Squared Error (MSE) loss: $\mathcal{L}_{full} = \mathbb{E}_D\big[||f(x)-l^*||^2 \big]$, 
where $\mathcal{L}_{full}$ is the loss for bowl fullness, $f$ is the feature extractor, $x$ is an RGB image, $l^*$ is the ground truth bowl fullness, which we manually label as values such as 0.8 or 0.5.

The overall loss for the representation learning is the following: $\mathcal{L}_{z} = \lambda_{CE} \mathcal{L}_{CE} + \lambda_{tri} \mathcal{L}_{tri} + \lambda_{temp} \mathcal{L}_{temp} + \lambda_{full} \mathcal{L}_{full}$, 
where $\lambda_{CE}$, $\lambda_{tri}$, $\lambda_{temp}$, and $\lambda_{full}$ are the coefficients for visual, physical, temporal, and geometric representation loss, respectively.

\subsection{Behavior Cloning}
As mentioned earlier, the primary objective of this paper is to learn a policy $\pi: \mathcal{S} \rightarrow \mathcal{A} $ that guides a robot's actions for food acquisition with BC. The BC loss is a Negative Log-Likelihood loss and is expressed as following: $\mathcal{L}=-\mathbb{E}_\mathcal{D} \big[\log \pi_\theta(a_t| s_{t-k:t}) \big]$, 
where $\theta$ is the parameters of the policy, $a_{t}$ is the expert action. Minimizing the BC loss $\mathcal{L}$ allows us to optimize the scooping actions for effective food manipulation. Unlike naive BC, we develop a richer and more informative multi-dimensional representation $z$ to enhance the robustness and generalizability of BC. We finetune $z$ during the training of BC with robot data. This representation enables the learned policy to tailor scooping strategies appropriately based on different contexts.

\section{Experiments} \label{sec:exp}
To validate our approach, we test it on a real robot. We design our experiments to answer the following key questions: (1) \textbf{Performance Comparison}: How well does our approach perform against other baselines for food manipulation, specifically in scooping tasks? (2) \textbf{Generalization Ability}: How well does our approach generalize to different contexts, including unseen food items with various colors and physical properties, and unseen bowls of different colors, shapes, and sizes? (3) \textbf{Advantage of Learned Representations}: How effectively does our approach leverage visual, temporal, and geometric learned representations for improved food manipulation? (4) \textbf{Critical Design Choices}: What design choices in our method are essential for achieving good performance?

\subsection{Experimental Setup} \label{sec:exp_setup}
We use a UR3e robot arm with RealSense cameras to caputure RGB images, and a spoon affixed to the arm for scooping tasks. To address question (1), we conduct evaluations in real-world tasks designed to scoop different types of food, such as granular cereals, liquid water, and semi-solid jello. These tasks provide quantitative comparisons between our approach and the baselines.  We use the following three evaluation metrics: \textbf{Success Rate (SUR)}: The ratio of successful scooping attempts (without spillage or collision) to the total number of scooping attempts. \textbf{Spillage or Failure Rate (SFR)}: The ratio of scooping attempts resulting in spillage or failure to the total number of scooping attempts. \textbf{Amount of Food Scooped (AFS)}: The quantity of food scooped after a certain number of sequential scooping attempts. 

To answer question (2), we perform zero-shot generalization testing on unseen bowls and food variants not encountered during robot demonstrations. We evaluate the performance of \ours~in these new contexts and compare it against the baseline methods. To address questions (3) and (4), we conduct ablation studies to validate our model's design and assess the effectiveness of visual, physical, temporal, and geometric representations. These studies help determine how each module contributes to overall performance. Additionally, we perform a qualitative analysis to evaluate the advantages of our learned visual and physical representations.

\subsection{Data Collection} \label{sec:data_collection}
We first collect a diverse set of labeled food images from the internet for representation learning. We then collect a set of robot demonstrations as training data for BC. The BC dataset consists of $N=30$ robot trajectories obtained through teleoperation using Gello \cite{wu2023gello}. Each trajectory includes pairs of RGB images and robot joint angles recorded throughout the scooping process. Specifically, we used a white circular bowl with volume 0.74 qt containing three types of food: granular cereals, semi-solid jello, and liquid water. Notably, we did not collect data for other bowls of different sizes, shapes and colors or food variants.

\subsection{Baselines}
We implemented several baselines for comparison: \textbf{Vis}: we used the method of Sundaresan et al. \cite{sundaresan2022learning} with only visual information, as our setup does not include haptic data. \textbf{SCONE} \cite{tai2023scone}: we applied this method with RGB images. \textbf{Pretrained Feature Extractors without Representation Learning}: we evaluated multiple methods using pretrained models to extract features, including: ImageNet pretrained ResNet-50 \cite{he2016deep}, DINO-v2 \cite{oquab2023dinov2}, MoCo-v3 \cite{chen2021empirical}. 

\begin{figure}[ht]
     \centering
     \begin{subfigure}[b]{0.49\linewidth}
         \centering
         \includegraphics[width=\textwidth]{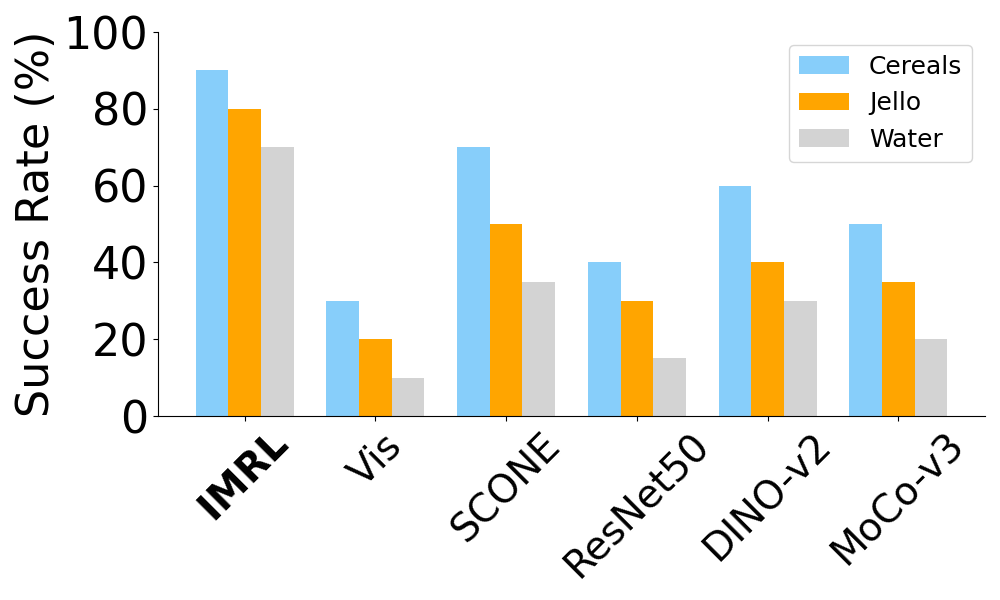}
         \caption{Success Rate ($\uparrow$) }
         \label{fig:sur}
     \end{subfigure}
     \hfill
     \begin{subfigure}[b]{0.49\linewidth}
         \centering
         \includegraphics[width=\textwidth]{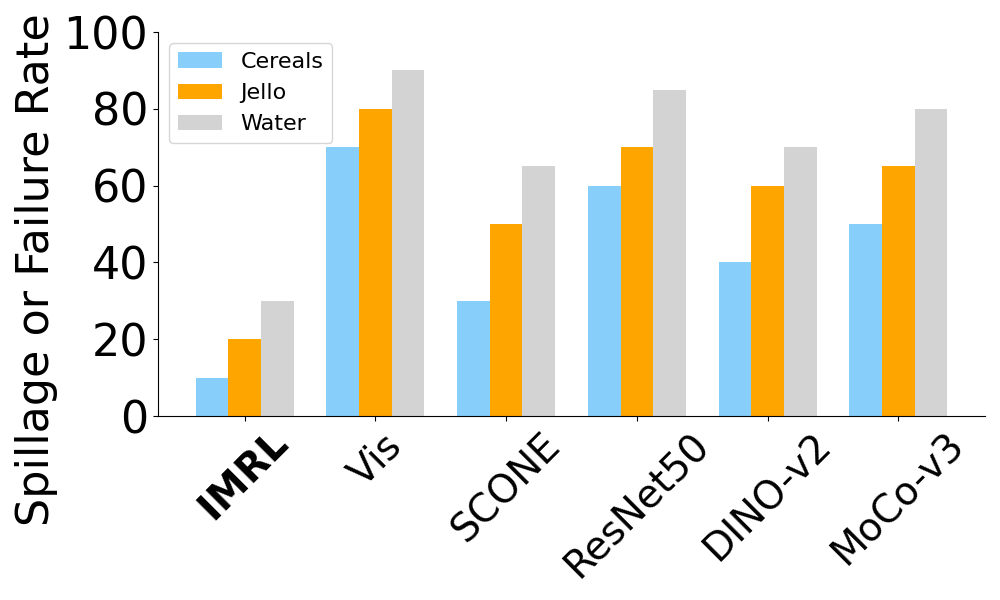}
         \caption{Spillage or Failure Rate ($\downarrow$)}
         \label{fig:spr}
     \end{subfigure}
     \medskip
     \begin{subfigure}[b]{0.49\linewidth}
         \centering
         \includegraphics[width=\textwidth]{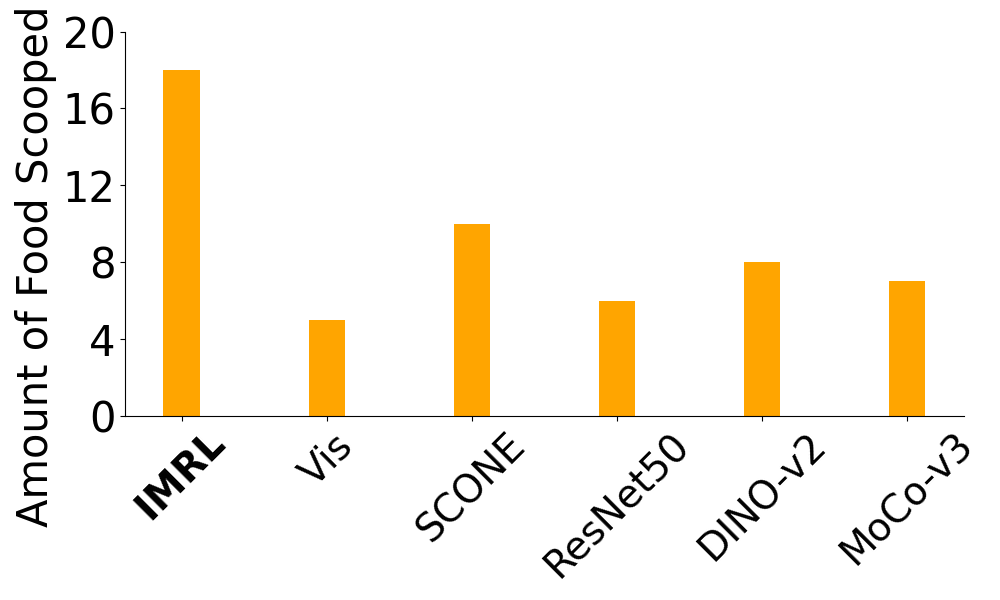}
         \caption{Amount of Food Scooped ($\uparrow$)}
         \label{fig:afs}
     \end{subfigure}
     \begin{subfigure}[b]{0.49\linewidth}
         \centering
         \includegraphics[width=\textwidth]{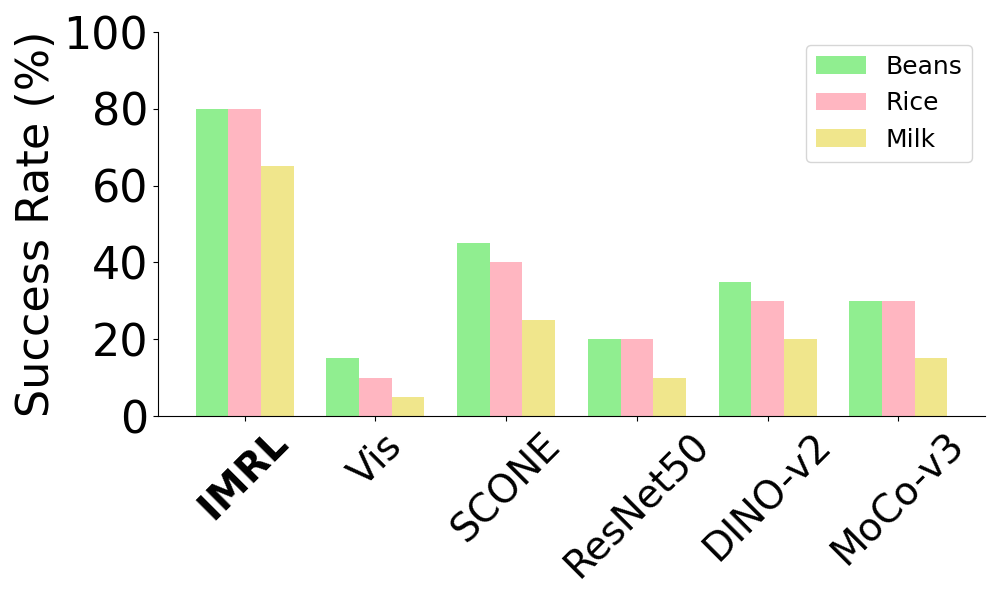}
         \caption{Generalization Testing}
         \label{fig:gen}
     \end{subfigure}
     \begin{subfigure}[b]{0.49\linewidth}
         \centering
         \includegraphics[width=\textwidth]{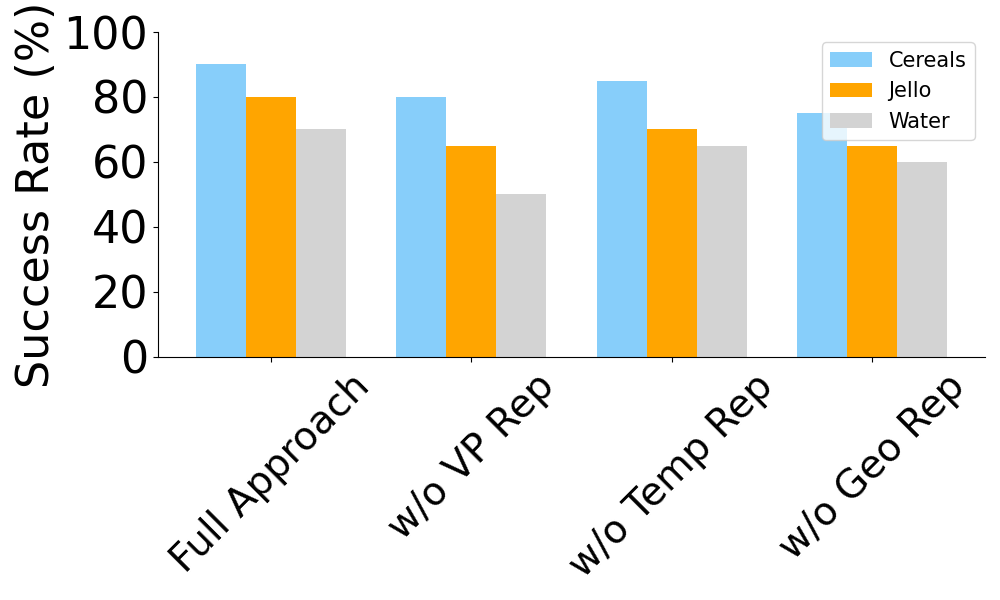}
         \caption{Ablation Studies}
         \label{fig:abl}
     \end{subfigure}
    \caption{\textbf{Experimental Results.} (a-c) Comparison of Success Rate, Spillage or Failure Rate, and Amount of Food Scooped between \ours~and other baselines. (d) Generalization testing on unseen scenarios, including unseen foods and bowls. (e) Ablation studies evaluating our full approach against three variants: without visual and physical representations, without temporal representation, and without geometric representation.}
    \label{fig:res}
\vspace{-10pt}
\end{figure}

\subsection{Experimental Results}


We first qualitatively analyze our learned visual and physical representations using 2D t-SNE visualization \cite{van2008visualizing} of representation embeddings for various food types. Fig. \ref{fig:pretrained} shows embeddings from a ResNet-50 model pretrained on ImageNet, which roughly separate food types but tend to cluster similar-colored items together, like black beans and jello, milk and yogurt. In contrast, as shown in Fig. \ref{fig:visual_only}, embeddings based solely on visual semantics provide better separation between food types compared to ResNet-50 but still ignore physical properties, like milk and yogurt, jello and soup. When incorporating both visual semantics and physical properties, our learned representations achieve better differentiation between food types and group items with similar physical properties, such as yogurt and jello; milk and water; beans, rice and cereals, as in Fig. \ref{fig:vp}. Foods with different physical properties are more distinctly separated, such as the mixture property of soup.

With the learned representations, we employ \ours~and compare it with other baselines on the real robot UR3 to answer question (1). As shown in Fig. \ref{fig:sur} and Fig. \ref{fig:spr}, \ours~achieves the highest success rate and the lowest spillage or failure rate across granular cereals, semi-solid jello, and liquid water. \ours~outperforms SCONE \cite{tai2023scone} by $20\%$, $30\%$, and $35\%$ success rate for cereals, jello, and water, respectively. Vis \cite{sundaresan2022learning} does not perform well  without haptic sensors. The performances of pretrained feature extractors without representation learning are worse than \ours~since they capture only visual information, lacking detailed physical properties, temporal dynamics, and geometric information of optimal scooping points and bowl fullness.

We also evaluate the quantity of food scooped after a certain number of scooping attempts. We scoop cereals for 10 sequential scooping actions. As shown in Fig. \ref{fig:afs}, \ours~successfully scoops $18\%$ of the total bowl volume (0.74 qt for the white circular bowl). While other methods fall short, due to spillage or failure.


\subsection{Generalization Testing}
We conduct zero-shot generalization testing on unseen scenarios from robot demonstrations to answer question (2). These scenarios involve various bowl types, including a large blue circular bowl, a small blue circular bowl, and a transparent square bowl, as well as different food items such as granular rice, black beans, green beans, yellow beans, and milk. To further assess the generalization capability of \ours~in handling foods with similar physical properties despite their differing color appearances, we mix black, green, and yellow beans and compare the success rate of \ours~with other baselines for acquiring unseen food. As shown in Fig. \ref{fig:gen}, \ours~performs comparably well on unseen foods, while the performance of other baselines decreases significantly, highlighting the limitation of standard BC methods in terms of robustness and generalizability.

\begin{figure}
    \centering
    \includegraphics[width=\linewidth]{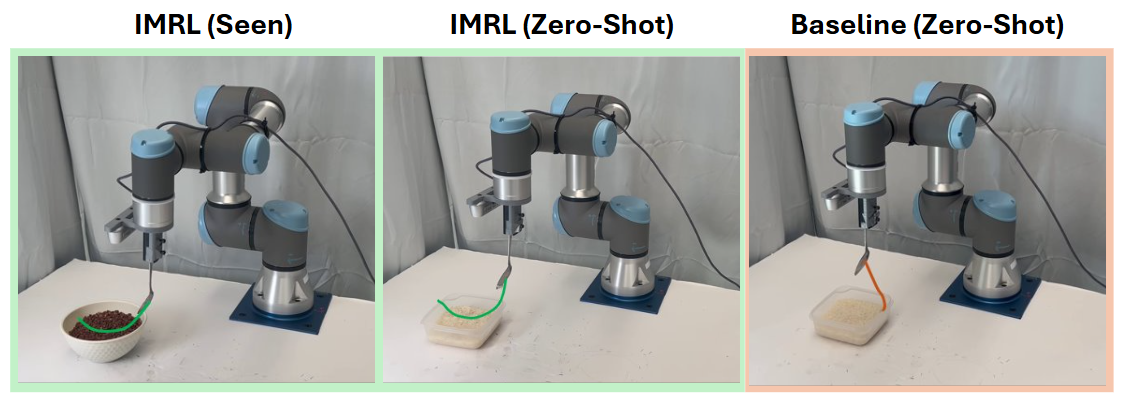}
    \caption{\textbf{Spoon trajectories of generalization testing.} \textbf{Left}: Spoon trajectory for scooping cereals from a white circular bowl (seen during training). \textbf{Middle}: Spoon trajectory for \ours~scooping unseen rice from an unseen transparent square bowl, showing similar motion patterns due to comparable granular properties. \textbf{Right}: Baseline trajectory using pretrained ResNet-50, which fails to scoop rice. Middle and Right use the same BC policy but just the representations are different.}
    \label{fig:traj}
    \vspace{-15pt}
\end{figure}

For qualitative comparison, we present the spoon trajectories from zero-shot generalization testing shown in Fig. \ref{fig:traj}. The left illustrates \ours~scooping cereals from a white circular bowl, a scenario seen during training. The middle shows \ours~handling unseen rice from an unseen transparent square bowl, demonstrating similar motion patterns due to the similar granular properties of the rice and cereals. On the right, the baseline using pretrained ResNet-50 fails to scoop rice, as it lacks the ability to capture detailed representations that account for food's physical properties. The middle and right examples utilize the same BC policy but differ in their representations.

\subsection{Ablation Studies}
We conduct ablation studies to address questions (3-4) and validate our model’s design by assessing the contributions of visual, physical, temporal, and geometric representations. We compare the performance of \ours~full approach with three ablated versions: one without visual and physical representations, one without temporal representation, and one without geometric representation. The results in Fig. \ref{fig:abl} demonstrate that removing any of these components leads to a decrease in success rate, confirming the effectiveness of each representation in enhancing overall performance.

\section{Conclusions} \label{sec:conclusion}
In this paper, we introduce \ours~(Integrated Multi-Dimensional Representation Learning), a novel approach to enhance the robustness and generalizability of behavior cloning (BC) for food acquisition in robotic assistive feeding. By integrating visual, physical, temporal, and geometric representations, \ours~enables adaptive food acquisition across diverse foods and bowl configurations, addressing limitations in existing methods that rely on surface-level geometric information. Our approach showes a $35\%$ improvement in success rate over the best-performing baseline, highlighting its effectiveness, including zero-shot generalization to unseen food and bowl combinations.


\textbf{Limitations and Future Work.} Despite the advantages of \ours, there are some limitations. We evaluate the quantity of food scooped after a certain number of scooping attempts; however, maximizing the amount of food scooped is not necessarily ideal, as individual preferences vary. Future work could explore conditioning the learned policy on a desired amount of food. Additionally, incorporating multimodal sensory inputs, such as force feedback or depth sensors, could capture more nuanced physical interactions between the robot and food items.

\bibliographystyle{ieeetr}
\bibliography{references}
\end{document}